\documentclass[runningheads]{llncs}
\usepackage{graphicx}
\usepackage{amsmath,amssymb} 
\usepackage{color}

\usepackage{url}
\usepackage[table]{xcolor}
\usepackage{bbm}
\usepackage{booktabs}
\usepackage{array}
\usepackage{epsfig}
\usepackage{dsfont}
\usepackage{multirow}
\usepackage{multicol,lipsum}

\usepackage{times}
\usepackage{helvet}
\usepackage{courier}
\usepackage[width=122mm,left=12mm,paperwidth=146mm,height=193mm,top=12mm,paperheight=217mm]{geometry}
\begin{document}
\title{A Network Structure to Explicitly Reduce Confusion Errors in Semantic Segmentation}

\titlerunning{Reduce Confusion Errors in Semantic Segmentation}
%
\author{Qichuan Geng\inst{1,2} \and
Xinyu Huang\inst{1} \and
Zhong Zhou\inst{2} \and
Ruigang Yang\inst{1}}
%
\authorrunning{Geng \textit{et al.}}
%

\institute{Baidu Research, Beijing, China, 100193\\
\email{\{gengqichuan,huangxinyu01,yangruigang\}@baidu.com}\\
 \and
School of Computer Science and Engineering, Beihang University, Beijing, China, 100083\\
\email{zz@buaa.edu.cn}}
\maketitle              
\begin{abstract}
Confusing classes that are ubiquitous in real world often degrade performance for many vision related applications like object detection, classification, and segmentation. The confusion errors are not only caused by similar visual patterns but also amplified by various factors during the training of our designed models, such as reduced feature resolution in the encoding process or imbalanced data distributions. A large amount of deep learning based network structures has been proposed in recent years to deal with these individual factors and improve network performance. However, to our knowledge, no existing work in semantic image segmentation is designed to tackle confusion errors explicitly. In this paper, we present a novel and general network structure that reduces confusion errors in more direct manner and apply the network for semantic segmentation. There are two major contributions in our network structure: 1) We ensemble subnets with heterogeneous output spaces based on the discriminative confusing groups. The training for each subnet can distinguish confusing classes within the group without affecting unrelated classes outside the group. 2) We propose an improved cross-entropy loss function that maximizes the probability assigned to the correct class and penalizes the probabilities assigned to the confusing classes at the same time. Our network structure is a general structure and can be easily adapted to any other networks to further reduce confusion errors. Without any changes in the feature encoder and post-processing steps, our experiments demonstrate consistent and significant improvements on different baseline models on Cityscapes~\cite{cordts2016cityscapes} and PASCAL VOC datasets~\cite{everingham2015pascal} (\textit{e.g.}, 3.05\% over ResNet-101~\cite{he2016deep} and 1.30\% over ResNet-38~\cite{wu2016wider}).
\end{abstract}
\section{Introduction}\label{sec:intro}
It is easy to find confusing classes that share similar visual patterns in real world. For instance, in a street view, road and sidewalk could have close color appearances. In an action video clip, hand clapping and boxing could share common moving patterns~\cite{liu2017easy}. In general, it could be impossible to completely avoid confusion errors even for human beings. However, confusion errors could be propagated and magnified throughout the training process in our models designed for various vision related tasks. In this work, we would like to mainly focus on the semantic segmentation task based on deep learning techniques. The extension could be made easily to other similar tasks such as object detection and image classification.

In semantic segmentation, a large amount of network structures has been proposed recently to deal with individual factors that could generate confusion errors. These factors include imbalanced data distributions and reduced resolution in the feature encoding process. Re-sampling and re-weighting (\textit{i.e.}, cost-sensitive) strategies~\cite{bunkhumpornpat2009safe,han2005borderline,jeatrakul2010classification,huang2016learning,bulo2017loss} are commonly applied to deal with imbalanced class distributions. However, performance may not be always improved due to some negative factors such as the over-fitting risk and increased dataset complexities when more minority classes are added. For instance, in Cityscapes dataset, over 90\% annotations come from the six majority classes including road, building, and vegetation. Less than 10\% annotations come from the remaining 13 minority classes.

In order to tackle reduced resolution problem caused by the feature encoder, a large amount of deep networks have been proposed based on the Fully Convolutional Neural Network (FCN)~\cite{sermanet2013overfeat,long2015fully}. Image pyramids could be fed into the same model and feature maps are fused together at the end~\cite{farabet2013learning,eigen2015predicting,pinheiro2014recurrent,lin2016efficient,chen2016attention,Chen2016DeepLab}. Multiple levels of decoders could be added to restore the details of feature maps~\cite{long2015fully,noh2015learning,ronneberger2015u,badrinarayanan2017segnet,lin2016refinenet,pohlen2017full,islam2017gated,wojna2017devil}. Novel convolution layers and pooling layers could be applied to capture multi-scale context information~\cite{Chen2016DeepLab,chen2017rethinking,chen2018encoder,zhao2017pyramid}. Current trends in semantic segmentation show that large kernels (\textit{e.g.}, $15\times15$ in~\cite{peng2017large}) or multiple atrous convolutions with different rates (\textit{e.g.}, (6, 12, 18) in~\cite{chen2017rethinking,chen2018encoder}) to capture much richer context information. However, it remains unclear how many different rates or kernel sizes should be selected and whether the selected ones are optimal to cover a large range of object sizes. In street views, the object sizes could be significantly different, \textit{e.g.}, a passing truck could have the same height of the image while a traffic light may have only dozens of pixels. As a result, even the large kernel mentioned in~\cite{peng2017large} may not big enough to cover a object like a truck and capture enough context information.

In this paper, we propose a novel network structure that is aimed to reduce semantic confusion errors explicitly. Comparing with existing methods, our network structure is able to deal with all the factors in a more direct manner. Moreover, this structure is general and can be easily integrated into any existing networks to further improve performance. Specifically, our proposed network structure mainly has two following contributions.
\begin{itemize}
  \item We propose a method to build and ensemble multiple subnets with heterogeneous output spaces. These subnets are built based on the discriminative confusing groups inferred from the normalized confusion matrix. Each subnet is aimed to enlarge the distances among the confusing classes within each confusing group without affecting unrelated classes outside the group.
  \item We propose an improved multi-class cross-entropy loss that considers both correct and incorrect labels. By adding a new term for the incorrect labels, both false negatives and false positives that are often caused by confusing classes are penalized directly. A re-weighting based on the confusion matrix is also applied for the new loss to further strengthen the penalization.
\end{itemize}

In Section~\ref{sec:related}, we introduce the related work in the semantic image segmentation. Section~\ref{sec:alg} describes our network structure and new loss function in details. An analysis of the loss function based on the information theory is given in this section. In Section~\ref{sec:exp}, we provide a set of experiments on the Cityscapes and Pascal VOC datasets. We give a conclusion in Section~\ref{sec:conc}.

\section{Related Work}\label{sec:related}
As mentioned in Section~\ref{sec:intro}, confusion could be magnified by various factors during the network training. We divide the factors into two categories, imbalanced data distribution and reduced feature resolution. In this section, we mainly describes related work to deal with these two categories.

\bigskip
\noindent\textbf{Imbalanced Data Distribution.} One method is to over-sampling the minority classes and/or under-sampling the majority classes. As this strategy changes the data distribution, over-sampling may result in over-fitting and under-sampling may remove possible valuable information. SMOTE and its variants~\cite{bunkhumpornpat2009safe,han2005borderline,jeatrakul2010classification} have been proposed to avoid the over-fitting by generating new non-replicated examples. Another direction, re-weighting method, imposes additional penalties on the minority classes without changing the data distribution. For instance, inverse frequency and median frequency re-weighting~\cite{caesar2015joint,mostajabi2015feedforward,xu2014tell,xu2015learning,eigen2015predicting} have been applied in the semantic segmentation works. In~\cite{shrivastava2016training}, online hard example mining (OHEM) is proposed to automatically select hard examples for training region-based ConvNet detectors. Huang \textit{et al.}~\cite{huang2016learning} formulates a new quintuplet sampling method and the triple-header loss for the large-scale imbalanced classification. A loss max-pooling layer~\cite{bulo2017loss} defines a new loss function that takes the highest loss from a pixel-level weighting function. This loss function could obtain performance gain over many minority classes (the performance on one minority class, ``truck", degrades for unknown reasons) in the Cityscapes dataset.

\bigskip
\noindent\textbf{Reduced Feature Resolution.} During the feature encoding, resolutions of feature maps gradually reduced in order to capture long range information that is less sensitive to the input image transformation. However, details of context are gradually compressed or lost in this encoding process. Long \textit{et al.} proposed the Fully Convolutional Network (FCN) for semantic segmentation in~\cite{long2015fully} that converts the fully-connected layers into convolutional layers in order to generate spatial label map directly. Based on this structure, a number of deep networks have been proposed. There are mainly three directions to improve the FCN. 1) Decoders could be added to gradually restore the context details. DeconvNet and SegNet~\cite{noh2015learning,badrinarayanan2017segnet} apply the inverse pooling layers to build glass-like networks to upsample feature maps. 2) The dilated convolution, also called atrous convolution~\cite{Chen2016DeepLab}, could be used to generate feature maps with higher resolutions during the encoding process. Due to limited GPU memory and other reasons, we still need to downsample the feature maps (typically $8\times$) in many networks. 3) Recent works focus on capturing multi-scale context information. In~\cite{peng2017large}, Peng \textit{et al.} integrate global convolutional networks (GCN) into different levels of feature maps and apply the deconvolution operations to restore high-resolution label maps. Large kernel (\textit{e.g.}, $15\times15$) used in the GCN enlarges the valid receptive field significantly. Chen \textit{et al.} proposed the atrous spatial pyramid pooling (ASPP) module in~\cite{chen2017rethinking} that arranges the atrous convolution operations in parallel with different atrous rates to obtain multi-scale context information. This module is further combined with a simple decoder module in~\cite{chen2018encoder}. 4) Other context modules (\textit{e.g.}, Conditional Random Fields (CRF)) also could be used as a post-processing step  or jointly trained with deep networks~\cite{krahenbuhl2011efficient,zheng2015conditional,lin2016efficient,schwing2015fully}.

\bigskip
\noindent\textbf{Batch Normalization.}
Batch normalization layer is a common-used layer in the semantic segmentation. As the importance of this layer has been discovered recently, we briefly introduce the work in the area although it is not connected with our contributions directly. Batch normalization parameters have been added to the ASPP module and found important during the training~\cite{chen2017rethinking,chen2018encoder}. The strategy is to compute batch normalization parameters with larger batch size and smaller feature map (\textit{e.g.}, $16\times$ downsampling rate) and freeze the parameters with larger feature map (\textit{e.g.}, $8\times$ downsampling rate). In~\cite{bulo2017place}, an in-place activated batch normalization (INPLACE-ABN) has been proposed to reduce the training memory so that batch size could be increased and statistics from the batch normalization could be more accurate. This novel batch normalization layer could boost the performance of the ResNet-38 model from 78.08\% to 79.40\% without other modifications of the network. Currently, the INPLACE-ABN ranks top one in the Cityscapes benchmark.

\bigskip
\noindent
In this section, we describe advanced deep networks that have greatly boosted the performance in the semantic image segmentation. So far, to our knowledge, none of existing works in this area has explored the reduction of confusion errors explicitly as proposed in this paper.

\section{Our Approach}\label{sec:alg}
\begin{figure}
  \centering
  \includegraphics[width=\linewidth]{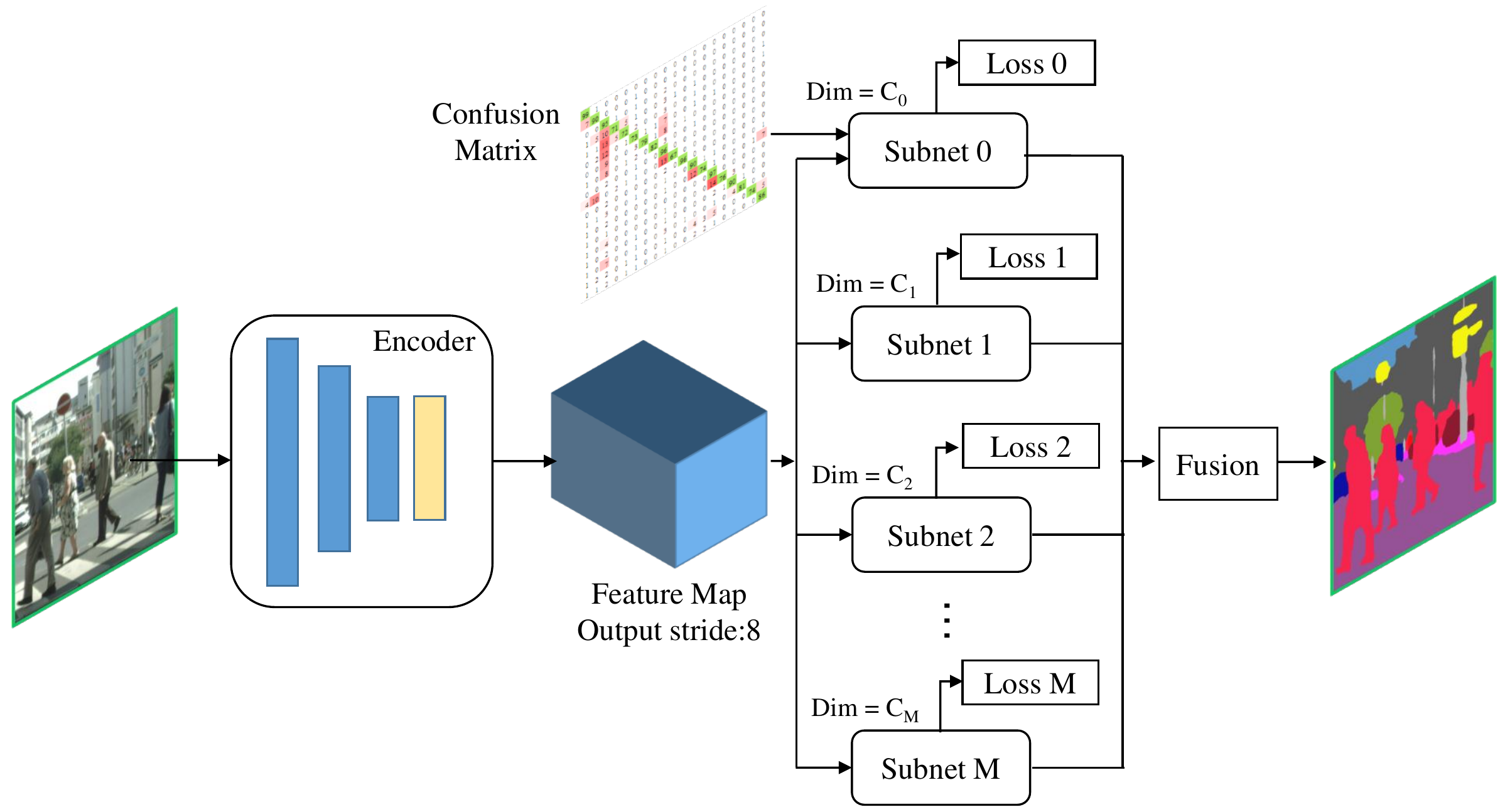}
  \caption{Overview of our proposed network structure that contains $M$ subnets.}\label{fig:pipeline}
\end{figure}

The overall network structure is shown in Figure~\ref{fig:pipeline}. Our subnets and loss layers could be easily integrated into most of existing network structures. We first separate the original network into two parts. The main part is used as the feature encoder. One or several convolutional layers and related batch normalization or activation layers used as our first subnet (\textit{i.e.}, subnet 0 in the Figure~\ref{fig:pipeline}). The number of remaining subnets ($M-1$) is determined by the number of the confusing groups. The number is not too large in general, such as three for the Cityscapes dataset that contains 19 classes. Moreover, in case that we have a very complex dataset, we can use a threshold to reduce the number so that selected subnets focus on more confusing groups. Each subnet is trained separately for each confusing group. After training, the heterogeneous output scores are transformed and fused together to obtain final probabilities or scores. Note that the structure of each subnet could be adjusted for a specific confusing group. For example, we could use more or less convolution layers, different atrous rates, and concatenated feature maps from different ResNet blocks.

\subsection{Discriminative Confusing Groups}
From the perspective of ensemble learning, we could view both fusion of subnets in our network and the ASPP module in~\cite{chen2017rethinking,chen2018encoder} as ensemble classifiers. Each subnet could be considered as a classifier in our network, and each atrous convolution operation followed by batch normalization in the ASPP module also could be viewed as a classifier. The major difference is that the ASPP module takes all the classes into the consideration although the module uses different atrous rates. It remains unclear whether the individual classifiers are diverse enough and how many component classifiers should be included in the ensemble.

\begin{figure}
  \centering
  \includegraphics[width=0.46\linewidth]{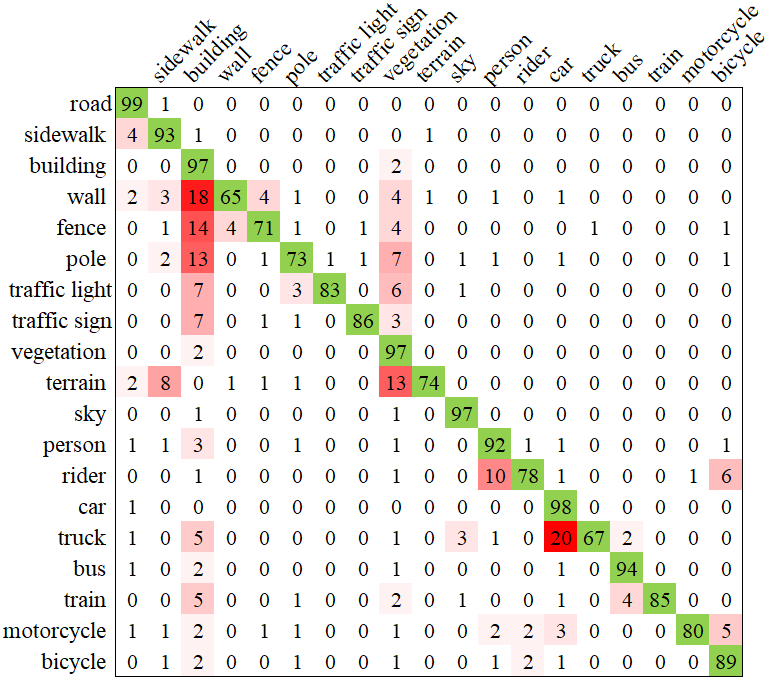}\quad
  \includegraphics[width=0.46\linewidth]{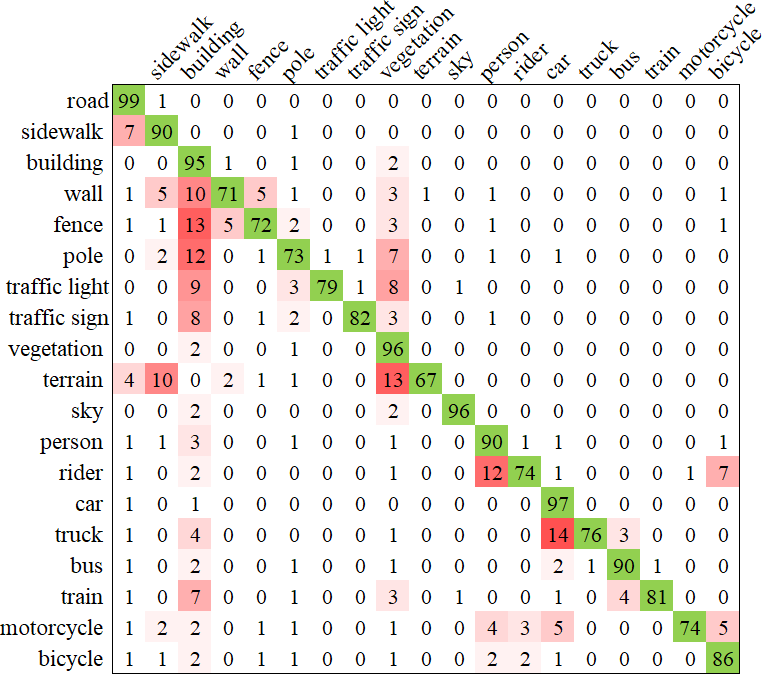}\\
  \caption{Two confusion matrices computed based on the pre-trained models of ResNet-101 (\textit{right}) and ResNet-38 (\textit{left}). The matrices share a very similar pattern that indicates the indepedent confusing groups.}\label{fig:confm}
\end{figure}

In Figure~\ref{fig:confm}, we compare the normalized confusion matrices computed from pre-trained models of ResNet-101 and ResNet-38 for the Cityscapes dataset. It is easy to find that computed matrices share a very similar pattern, such as the ``wall" class is strongly related to the ``building" class in both networks although misclassification rates are 18\% and 10\% respectively. Given the pattern, we can divide all the classes into few discriminative confusing groups where the inter-group confusing errors are very small and could be neglected. As a result, the number of discriminative confusing groups might be a practical and appropriate ensemble size or ensemble cardinality.

\subsection{Improved Cross-Entropy Loss}
In information theory, the cross-entropy between the ground truth distribution $p$ (i.e., the one-hot label in classification) and the estimated distribution $q$ is given by
\begin{equation}\label{eq:ce}
  H(p,q) = -\sum_{j=1}^K p_j\log q_j
\end{equation}
where $K$ is the number of classes and $q=e^{\phi(x)}/\sum_j e^{\phi_j(x)}$ is the output of softmax classifier. $\phi$ represents the network and $x$ is the input image. This term also can be interpreted as the loss associated with the probability assigned to the correct class without considering the relation between the correct class and the remaining classes, especially the confusing classes. In order to reduce the confusion to another incorrect class, intuitively, we also need to reduce the probability assigned to the incorrect class. As a result, we can formulate a new loss given by
\begin{equation}\label{eq:ce1}
  L = H(p,q) - H(1-p,1-q)
\end{equation}
 In above equation, we treat the correct class and the remaining classes equally in which confusing classes are still not taken into consideration. Hence, we weight the new loss using a weight matrix ($C=[c_{ij}]_{K\times K}$) that could be computed from the normalized confusion matrix. The equation~\ref{eq:ce1} is then converted into
 \begin{equation}\label{eq:ce2}
   L=-\sum_{i=1}^K c_{ii}p_i\log q_i+\lambda \sum_{i=1}^K\sum_{j=1}^K c_{ij} p'_j\log(1-q_j)
 \end{equation}
 where $p'=1-p$ and $\lambda$ is used to balance the losses between correct classes and incorrect classes. The derivative of the loss function~\ref{eq:ce2} with respect to $\phi(x)$ is
 \begin{align}\label{eq:deri}
   \frac{\partial L}{\partial \phi_k} &=-c_{ii}\frac{1}{q_i}\frac{\partial q_i}{\partial \phi_k}-\lambda\sum_{j=1,j\neq i}^K c_{ij} \frac{1}{q_j-1}\frac{\partial q_j}{\partial \phi_k} \nonumber\\
   &=q_k(c_{ii}+\lambda\sum_{j=1,j\neq i}^K c_{ij}\frac{q_j}{q_j-1})-c_{ii}, k=i \nonumber\\
   &=q_k(c_{ii}+\lambda\sum_{j=1,j\neq i}^K c_{ij}\frac{q_j}{q_j-1})-\lambda\frac{q_k}{q_k-1}c_{ik}, k\neq i
 \end{align}

 \subsection{Fusion of Heterogeneous Output Spaces}
As subnets belong to heterogeneous output spaces (denoted as source output spaces), we also need to find a way to transform them first into the target output space that contains all the labels and then we can apply traditional ensemble methods such as sum rule or product rule~\cite{kittler1998combining}. There are a number of ways to do the transformation, such as the regression model~\cite{shi2010transfer,shi2010predictive} and neural network~\cite{cisse2016adios}. In~\cite{shi2010predictive}, the \textit{similarity preserving principle} states that, for every pair of classes $(v_1,v_2)$ in the source output space, there is a similarity indicator $\pi$ so that $\pi(v_1,v_2)=\pi(f(v_1),f(v_2))$, where $f$ is the transformation function from the source output space to the target output space.

\begin{figure}
  \centering
  \includegraphics[width=0.9\linewidth]{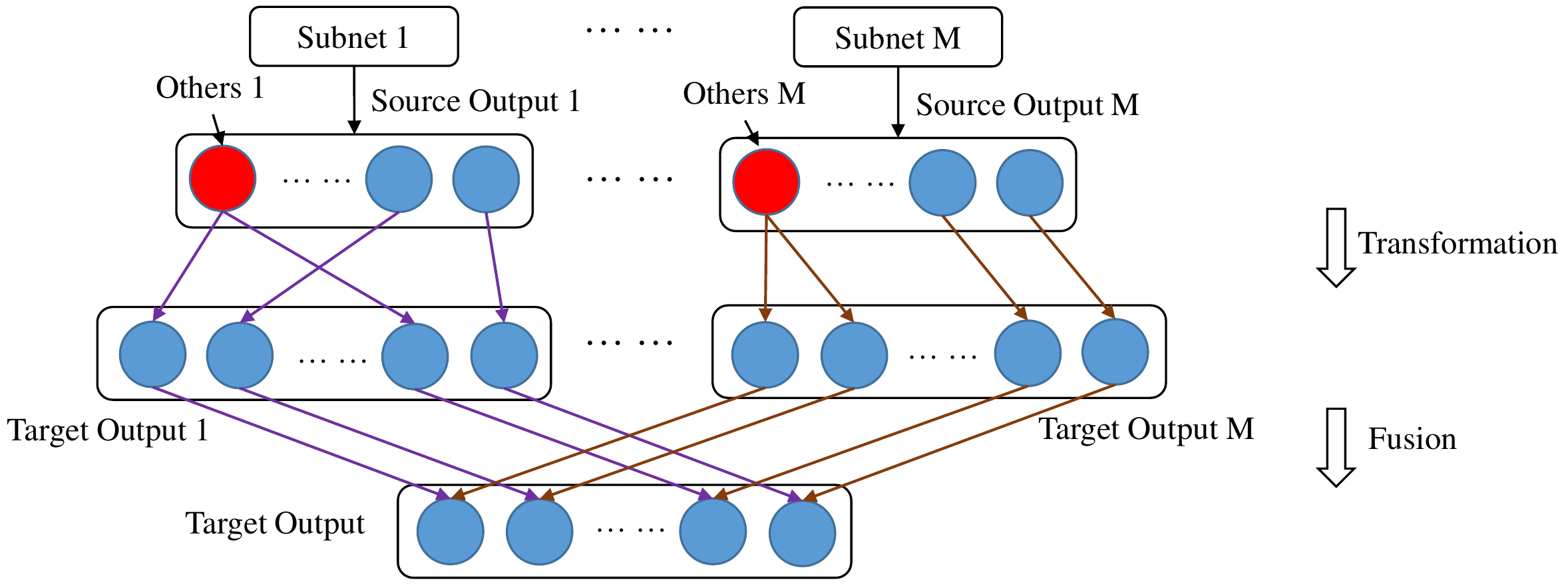}
  \caption{The transformation from source output space to target output space that satisfies \textit{similarity preserving principle} in~\cite{shi2010predictive}.}\label{fig:space}
\end{figure}

In each subnet in our network, we have classes belong to the corresponding confusing group and one ``others" class that includes all the classes outside the group. As the source output spaces have certain overlapping areas, we do not need to apply the regression model in~\cite{shi2010predictive,shi2010transfer}. Here we could apply a straightforward method as illustrated in Figure~\ref{fig:space}. For confusing classes in the group, there is one-to-one mapping between source output space and target output space. For the ``others" class, we build a one-to-many mapping from the source to all the remaining classes outside the group in the target output space. It is not difficult to prove that this simple transformation satisfies the \textit{similarity preserving principle}.

It is also possible design another network to learn the transformation, which might be one of our future works. In the experiments, we demonstrate that, even with our simple transformation and fusion, we still can achieve consistent improvements over the baseline model.


\section{Experiments}\label{sec:exp}
We evaluate our network structure on Cityscapes dataset~\cite{cordts2016cityscapes} and the extended PASCAL VOC dataset~\cite{everingham2015pascal} using the official MXNET tool~\cite{chen2015mxnet}. The performance is evaluated based on the mean intersection-over-union (mIoU). During the training, we use the standard SGD~\cite{krizhevsky2012imagenet} with momentum 0.9 and weight decay 0.0005. The initial learning rate is 0.002 and updated in a linear schedule. Data augmentation such as mean subtraction, random crop, and random left-right flipping, are applied during the training.

\subsection{Cityscapes}
Cityscapes contains 24,998 street views collected in 50 cities. 5,000 images with resolution $2048\times1024$ are fine annotated and remaining 19,998 are coarsely annotated. The 5,000 fine annotated images are further divided into train, validation, and test sets that have 2,975, 500, and 1,525 images, respectively. 19 semantic object classes are used for evaluation.

Two baseline models, ResNet-101 and ResNet-38, are selected for the experiments on Cityscapes dataset. The last 1000-way classification layer of the original ResNet-101 is removed. The feature stride is reduced from 32 to 8 for the semantic segmentation task by changing the convolution strides for block 3 and block 4. ResNet-101 is pre-trained on ImageNet~\cite{russakovsky2015imagenet} and fine-tuned on Cityscapes for 100 epochs. In~\cite{wu2016wider}, a mIoU 73.63\% is reported on Cityscapes validation dataset using ResNet-101. Our ResNet-101 obtains 74.70\%, which is 1.03\% higher. The ResNet-38 baseline model is the original released model in~\cite{wu2016wider} with a mIoU 78.08\%.

\begin{figure}[t]
  \centering
  \includegraphics[width=\linewidth]{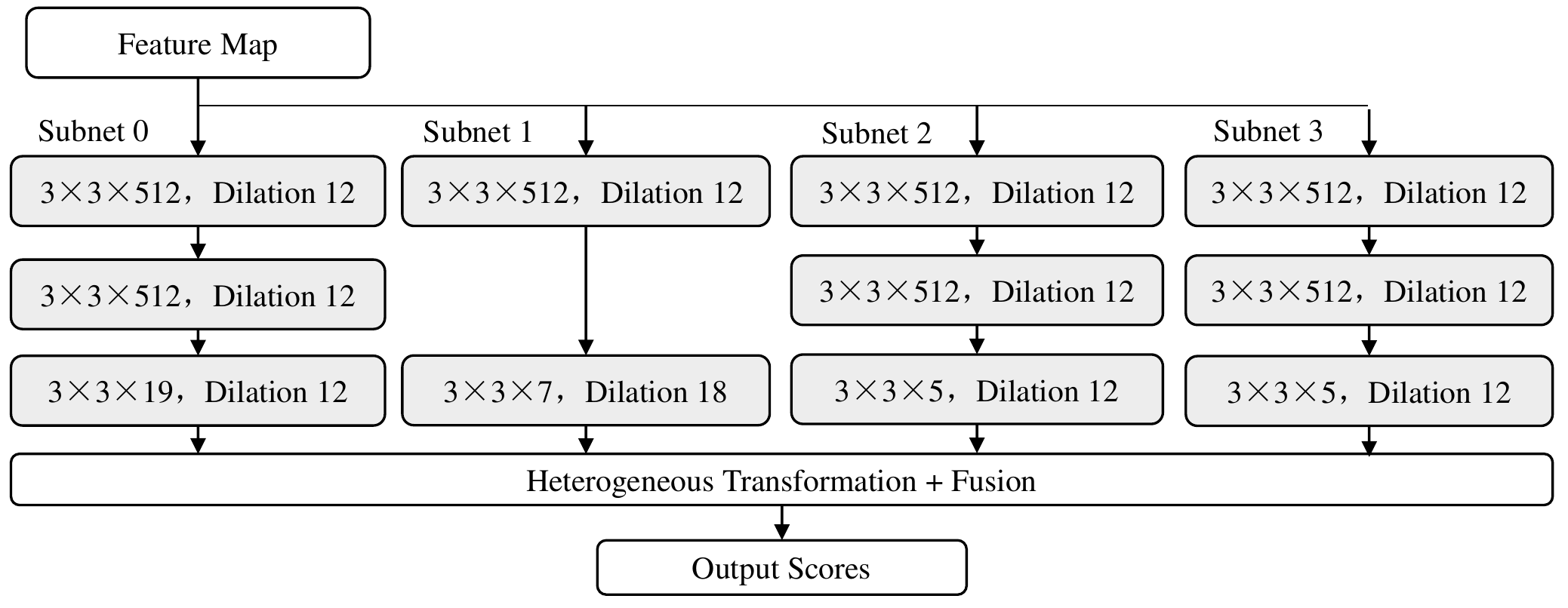}
  \caption{The structure and shape information of four subnets used for Cityscapes dataset.}\label{fig:csub}
\end{figure}

We partition all the classes into three confusing groups and build three additional subnets. Table~\ref{tb:group} gives us the details of these confusing groups. The ``others" class in each subnet contains all the remaining classes that are unrelated to the confusing classes within the corresponding group. Figure~\ref{fig:csub} shows the structure of all the subnets. All the parameters in the feature encoder are fixed during the training of subnets.

\begin{table}
  \centering
  \caption{Class names of three confusing groups on Cityscapes dataset.}\label{tb:group}
  \begin{tabular}{ p{8em}  p{8em}  p{22em}  }
     \toprule
     \multicolumn{1}{c}{subnet} & \multicolumn{1}{c}{nClass} & \multicolumn{1}{c}{Class Name} \\
     \midrule
     \multicolumn{1}{c}{subnet 0} & \multicolumn{1}{c}{19} & all the 19 classes \\
     \midrule
     \multicolumn{1}{c}{subnet 1} & \multicolumn{1}{c}{7} & others1, building, wall, fence, pole, traffic light, traffic sign \\
     \midrule
     \multicolumn{1}{c}{subnet 2} & \multicolumn{1}{c}{5} & others2, car, truck, bus, train \\
     \midrule
     \multicolumn{1}{c}{subnet 3} & \multicolumn{1}{c}{5} & others3, person, rider, motorcycle, bicycle \\
     \bottomrule
   \end{tabular}
\end{table}

\begin{figure}
  \centering
  \includegraphics[width=\linewidth]{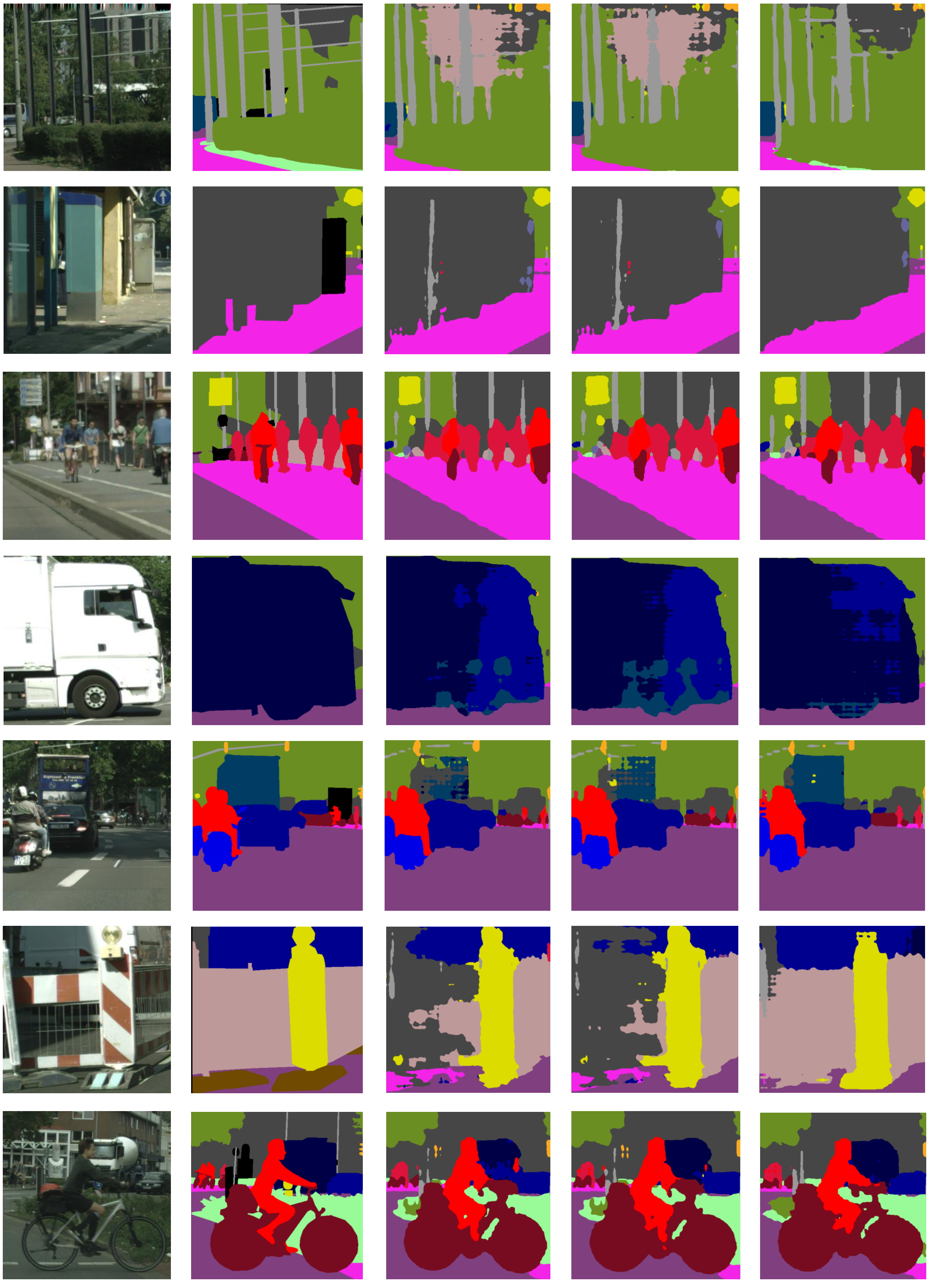}
  \caption{Examples of semantic segmentation results on Cityscapes (cropped for visualization purpose). For every row, we list the cropped input image, ground truth label, label estimated from the baseline, label estimated from baseline + subnets, and label estimated from baseline + subnets + improved loss. Confusing classes for each column are (from left to right): (building, pole), (building, pole, fence), (rider, person), (truck, bus), (truck, building), and (fence, building).}\label{fig:cimages}
\end{figure}

\textbf{Improvement on ResNet-101.} Table~\ref{tb:1011} shows the experiments using the subnets and improved cross-entropy loss. Without using our new loss, we improve the mIoU by 0.85\%. The improvement of our fusion methods mainly focus on the confusing classes. The average gain on these confusing groups are 0.45\%, 1.57\% and 1.70\%, respectively. In order to evaluate new loss, we only use subnet 0 by removing subnets from 1 to 3. We could improve mIoU to 76.21\% with $\lambda = 5$. When three subnets and new loss are stacked together, we are able to obtain mIoU 77.75\%, which is 3.05\% improvement over the baseline. Figure~\ref{fig:cimages} presents examples of visual results for some confusing classes (\textit{e.g.}, wall, pole, rider, truck, and fence).

\begin{figure}
  \centering
  \includegraphics[width=0.9\linewidth]{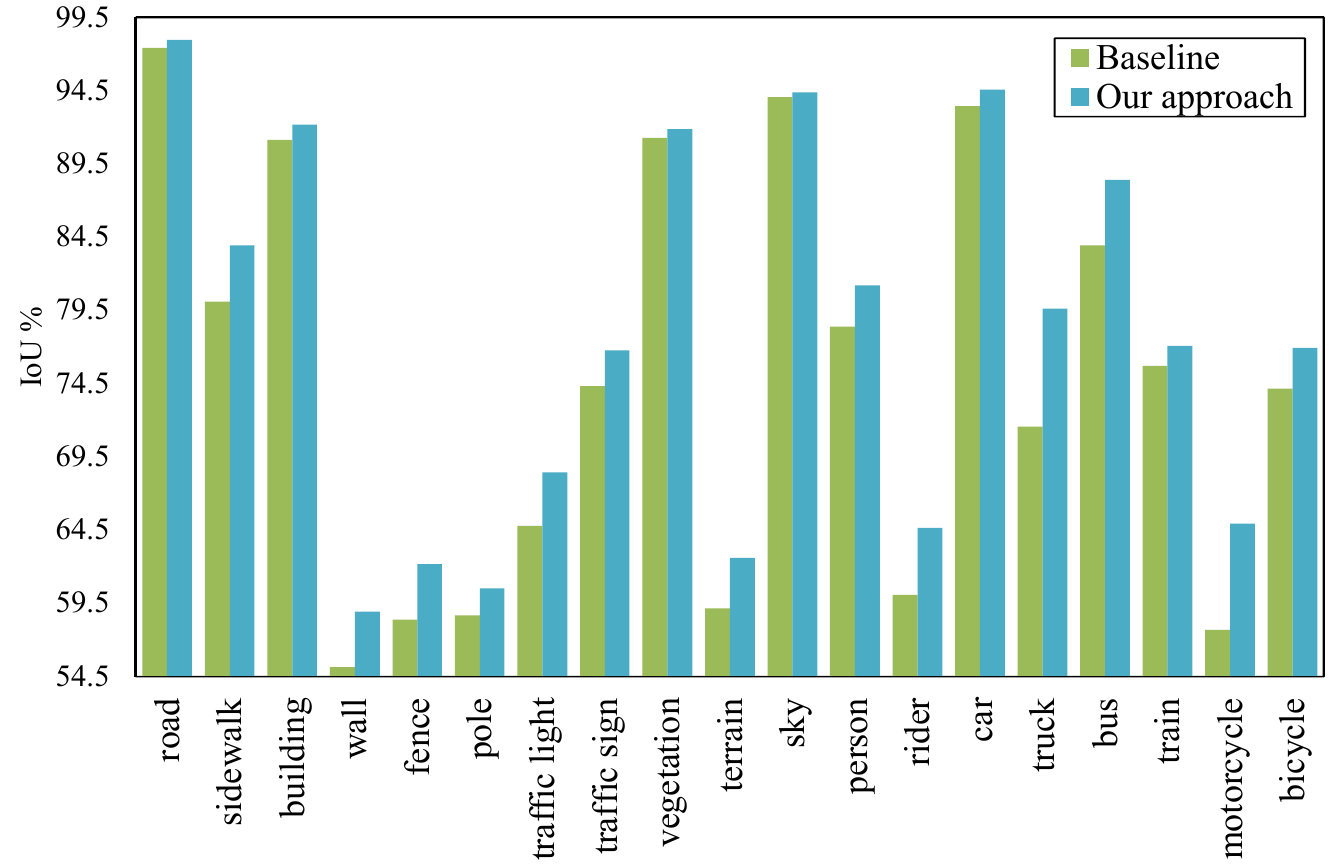}
  \caption{Comparison of per-class performance between our approach and the baseline.}\label{fig:iou_comp1}
\end{figure}

\begin{figure}
  \centering
  \includegraphics[width=0.9\linewidth]{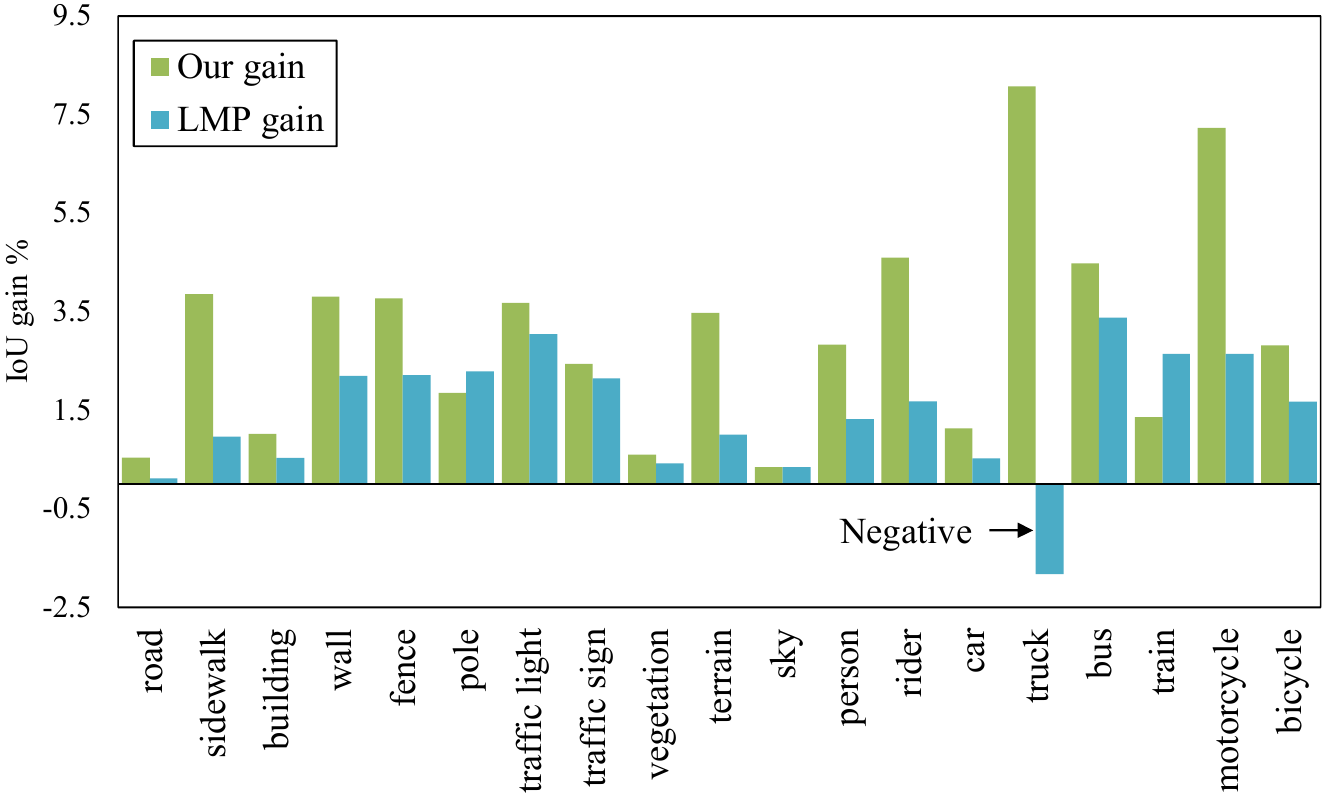}
  \caption{Comparison of per-class performance gain between our approach and the LMP~\cite{bulo2017loss}.}\label{fig:iou_comp2}
\end{figure}

\textbf{Per-class Performance on ResNet-101.}
In Figure~\ref{fig:iou_comp1}, we demonstrate the per-class performance of our approach versus the ResNet-101 baseline. We find that IoU values of 18 classes are improved greatly comparing with the baseline. Certain confusing classes, such as sidewalk, wall, fence, rider, truck, bus, and motorcycle, have IoU gains over 3.5\%. This result shows that our subnets and improved cross-entropy loss are effective to reduce confusion errors. In Figure~\ref{fig:iou_comp2}, we further compare our per-class IoU gains with the per-class IoU gains from LMP~\cite{bulo2017loss}. We find that our IoU gains for most of classes are larger than the gains from LMP. One possible reason is that the LMP is mainly designed to reduce imbalanced data problems. However, confusion errors could come from other factors not limited to the imbalanced data distribution. Therefore, this indicates that it could be more beneficial and effective to handle confusion errors explicitly.

\textbf{Comparison with Other Approaches.} Here we present a comparison between our approach and other approaches. In order to show a comprehensive comparison that includes baseline models, parameters, mIoU gains, we mainly choose the approaches that have been reported in published articles in recent years. We roughly partition these approaches into three different categories based on the problems being claimed to resolve by authors, i.e., ``I" for imbalanced data, ``F" for feature extraction (\textit{e.g.}, multi-scale context information), ``B" for improvement on batch normalization. Table~\ref{tb:comp} shows the comparison. As data augmentation has been done for all the approaches in the table, this option is not shown in the table.

Although the models in the table are aimed to resolve different problems, our improved mIoU and mIoU gain are comparable with the existing approaches proposed in recent years. Notice that I-ABN~\cite{bulo2017place} currently ranked first with mIoU 82.0\% in Cityscapes benchmark. If we only consider the modified batch normalization layer, the improved mIoU (77.58\%) reported in the paper is close to our improved mIoU (77.75\%) for baseline ResNeXt 101 and Resnet 101, respectively.

Based on this table, we find that, in order to obtain the optimal segmentation performance on Cityscapes, it is necessary to integrate multiple techniques together, such as ASPP, I-ABN, and LMP. This tells us the network structure should be general and flexible to fit into other different structures. Our approach is general, which can be easily combined with many existing works to further boost performance.

\begin{table}
  \centering
  \caption{Comprehensive comparison with other approaches on validation set of Cityscapes. The first column means the category of the problem being claimed to resolve by the authors. OS means output stride, MS means multi-scale input images used in testing. $\surd$ means the technique is applied in the model, a white space means it is not used in the model, ``-" means that the technique or the number is not reported in the article.}\label{tb:comp}
  \begin{tabular}{p{2em} c c c c c c c c c c}
     \toprule
       & model & OS & ASPP & CRF & MS & baseline & mIoU & mIoU & gain \\
       & & & & & & & (baseline) & (improved) & \\
     \midrule
     I & LMP~\cite{bulo2017loss} & 8 & $\surd$ & & & DeepLabV2 & 73.63 & 75.06 & 1.43 \\
     I & FCRNs~\cite{wu2016high} & 8 & & & & ResNet 101 & 68.58 & 71.16 & 2.58 \\
     I & FCRNs~\cite{wu2016high} & 8 & & & & ResNet 152 & 69.69 & 71.51 & 1.82 \\
     \midrule
      F & DeepLabV2~\cite{Chen2016DeepLab} & 16 & $\surd$ & & & ResNet 101 & 66.6 & 71.0 & 4.4 \\
      F & DeepLabV2~\cite{Chen2016DeepLab} & 16 & $\surd$ & $\surd$ & & ResNet 101 & 66.6 & 71.4 & 4.8 \\
      F & GCN~\cite{peng2017large} & 8 & & & & ResNet 152 & - & 76.9 & - \\
      F& GCN~\cite{peng2017large} & 8 & & $\surd$ & $\surd$ & ResNet 152 & - & 77.4 & -\\
      F& ResNet38~\cite{wu2016wider} & 8 & & & & ResNet 38& - & 77.86 & - \\
     \midrule
     B & I-ABN 101~\cite{bulo2017place} & 8 & & & & ResNeXt 101 & 74.42 & 77.58 & 3.16\\
     B & I-ABN 152~\cite{bulo2017place} & 8 & & & & I-ABN 101 & 77.58 & 78.49 & 0.91 \\
     \midrule
      & Ours & 8 & & & & ResNet 101 & 74.70 & 77.75 & 3.05\\
     \bottomrule
   \end{tabular}
\end{table}

Naturally we would like to apply our approach to existing state-of-the-art algorithms listed in Table~\ref{tb:comp}. However, after careful examinations, the released versions of these algorithms are not sufficient to allow modifications, usually only the testing models are released. As a result, we choose to evaluate our approach on ResNet-38 released model (mIoU is 78.08\% that is slightly higher than 77.86\% reported in the paper). Without using our new loss, we improve the mIoU by 0.99\%. With the new loss and subnet 0, we can improve mIoU to 79.38\%, which is 1.30\% improvement over the released model of ResNet-38.

\begin{table}

\parbox{.49\linewidth}{
\centering
  \caption{Improvement on ResNet-101.}\label{tb:1011}
  \begin{tabular}{ p{3em} p{3em} p{3em} p{12em} p{12em} }
     \toprule
     subnet & \multicolumn{1}{c}{0} & \multicolumn{1}{c}{0 - 3} & \multicolumn{1}{c}{0} & \multicolumn{1}{c}{0 - 3} \\
     \midrule
     loss & \multicolumn{1}{c}{CE} & \multicolumn{1}{p{3em}}{~~~CE} & \multicolumn{1}{p{4em}}{~~New CE} & \multicolumn{1}{p{4em}}{~~New CE} \\
     \midrule
     mIoU & \multicolumn{1}{c}{74.70} & \multicolumn{1}{c}{75.55} & \multicolumn{1}{c}{76.21} & \multicolumn{1}{c}{77.75} \\
     \bottomrule
   \end{tabular}
}
\hfill
\parbox{.49\linewidth}{
  \centering
  \caption{Improvement on ResNet-38.}\label{tb:r38}
  \begin{tabular}{ p{3em} p{3em} p{12em} p{5cm} }
     \toprule
     subnet & \multicolumn{1}{c}{0} & \multicolumn{1}{c}{0 - 3} & \multicolumn{1}{c}{0} \\
     \midrule
     loss & \multicolumn{1}{c}{CE} & \multicolumn{1}{p{5em}}{~~~~~CE} & \multicolumn{1}{p{5em}}{~~~New CE} \\
     \midrule
     mIoU & \multicolumn{1}{c}{78.08} & \multicolumn{1}{c}{79.07} & \multicolumn{1}{c}{79.38} \\
     \bottomrule
   \end{tabular}
}
\end{table}

\subsection{PASCAL VOC 2012}
PASCAL VOC 2012 has 1,464 images for training, 1,449 images for validation, and 1,456 images for testing. 21 object classes including the ``background" class are annotated. We also use the Semantic Boundaries dataset~\cite{hariharan2011semantic} as the auxiliary dataset, resulting in 10,582 images for training.

ResNet-101 is selected for the experiments on PASCAL VOC dataset. The structure of ResNet-101 is the same as the one used for evaluation of Cityscapes dataset. Similarly, ResNet-101 is pre-trained on ImageNet and fine-tuned on VOC dataset for 80 epochs. In~\cite{wu2016wider}, a mIoU 75.35\% is reported with ResNet-101 on PASCAL VOC validation dataset. Our ResNet-101 obtains 75.43\%, which is slightly higher.

\begin{figure}[t]
  \centering
  \includegraphics[width=0.47\linewidth]{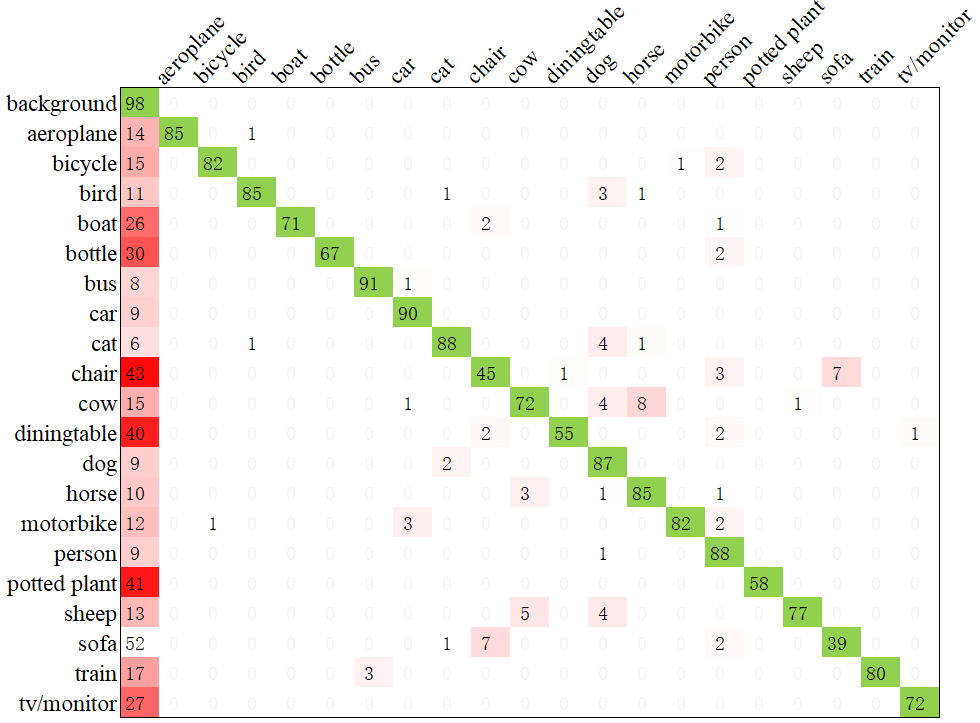}\quad
  \includegraphics[width=0.47\linewidth]{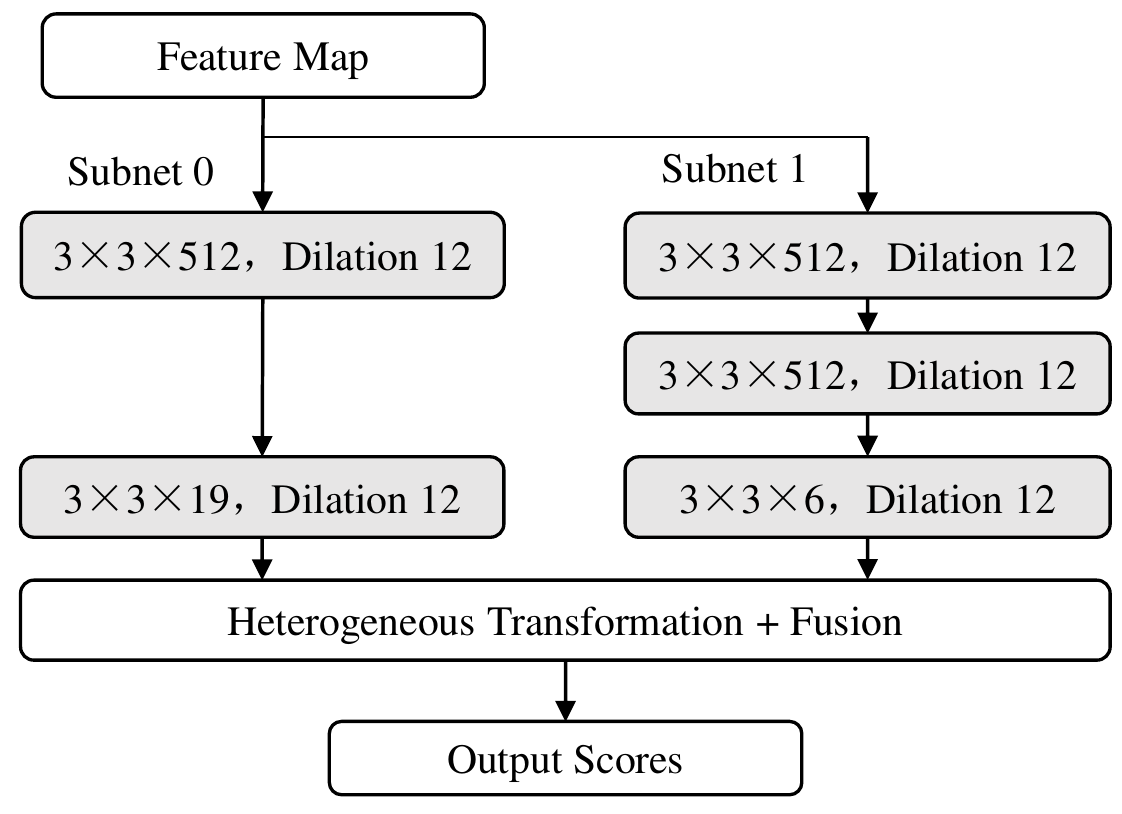}\\
  \caption{(\textit{left}) The confusion matrix for PASCAL VOC dataset. (\textit{right}) The structure and shape information of our subnets used for PASCAL VOC dataset.}\label{fig:confm3}
\end{figure}

Based on the confusion matrix shown in the left of Figure~\ref{fig:confm3}, we find that a number of classes are confused with the ``background" class. Therefore, only one subnet is added for this confusing group (\textit{i.e.}, others, background, chair, dining table, potted plant, and sofa). The structure of our network is shown in the right of Figure~\ref{fig:confm3}.

\begin{figure}
  \centering
  \includegraphics[width=\linewidth]{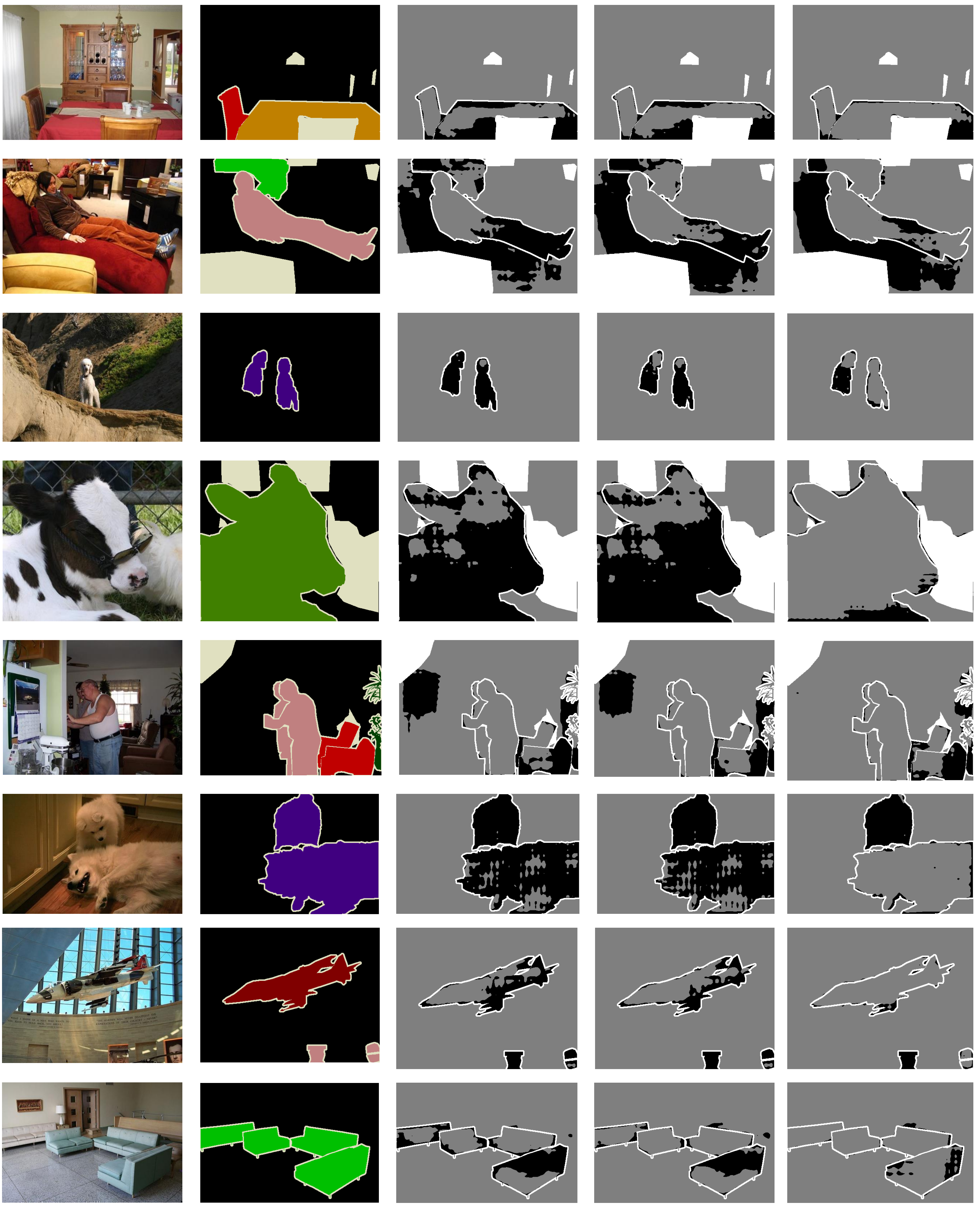}
  \caption{Examples of semantic segmentation results on PASCAL VOC dataset. For every row, we list input image, ground truth label, error map for baseline, error map for baseline + subnets, error map for baseline + improved loss. Error maps in the last three rows represent the label inconsistencies, in which black color means labels are different and gray color means labels are the same.}\label{fig:vimages}
\end{figure}

\textbf{Improvement on ResNet-101.}
Table~\ref{tb:1012} shows the experiments using the subnets and improved cross-entropy loss. mIoU is increased to 75.51\% when subnet 1 is used. mIoU is further increased to 76.91\% when the improved cross-entropy loss is applied. Some visual results are shown in Figure~\ref{fig:vimages}.

\begin{table}
  \centering
  \caption{Improvement on PASCAL VOC dataset using ResNet-101.}\label{tb:1012}
  \begin{tabular}{ p{4em} p{6em} p{6em} p{12em} }
     \toprule
     subnet & \multicolumn{1}{c}{0} & \multicolumn{1}{c}{0 and 1} & \multicolumn{1}{c}{0 and 1} \\
     \midrule
     loss & \multicolumn{1}{c}{CE} & \multicolumn{1}{c}{CE} & \multicolumn{1}{p{5em}}{~~~New CE}  \\
     \midrule
     mIoU & \multicolumn{1}{c}{75.43} & \multicolumn{1}{c}{76.51} & \multicolumn{1}{c}{76.91} \\
     \bottomrule
   \end{tabular}
\end{table}

\section{Conclusion}\label{sec:conc}
In this paper, we present a novel network structure to reduce semantic confusion errors that could come from different factors. While most existing works are designed to deal with individual factors, our approach is a more direct way to handle confusion errors. Our approach consists of two major components: 1) an ensemble of subnets with heterogeneous outputs from discriminative confusing groups estimated from the normalized confusion matrix; 2) an improved cross-entropy loss with a new term that penalizes both false negatives and false positives often caused by confusing classes. Our experiments show that both components are effective and improve segmentation performance over different baseline models and datasets with different complexities. More importantly, our approach is general and flexible, which can be easily fit into most of existing network structures.

\clearpage

\bibliographystyle{splncs04}
\bibliography{egbib}
\end{document}